\documentclass[lettersize,journal]{IEEEtran}
\usepackage{amsmath,amsfonts}
\usepackage{algorithmic}
\usepackage{algorithm}
\usepackage{array}
\usepackage[caption=false,font=normalsize,labelfont=sf,textfont=sf]{subfig}
\usepackage{textcomp}
\usepackage{stfloats}
\usepackage{url}
\usepackage{verbatim}
\usepackage{graphicx}
\usepackage{cite}
\usepackage{caption}
\usepackage{capt-of}
\usepackage{booktabs}
\usepackage{multirow}
\usepackage{amssymb}
\usepackage{xcolor}

\usepackage{xspace}
\makeatletter
\DeclareRobustCommand\onedot{\futurelet\@let@token\@onedot}
\def\@onedot{\ifx\@let@token.\else.\null\fi\xspace}

\def\eg{\emph{e.g}\onedot}

\makeatother

\begin{document}

\title{AvatarVTON: 4D Virtual Try-On for Animatable Avatars}

\author{
    Zicheng Jiang, 
    Jixin Gao,
    Shengfeng He,
    Xinzhe Li,
    Yulong Zheng,
    Zhaotong Yang,
    Junyu Dong,
    Yong Du
  
}


\markboth{}%
{Shell \MakeLowercase{\textit{et al.}}: A Sample Article Using IEEEtran.cls for IEEE Journals}


\twocolumn[{
\renewcommand\twocolumn[1][]{#1}
\maketitle
\begin{center}
  \includegraphics[width=\textwidth]{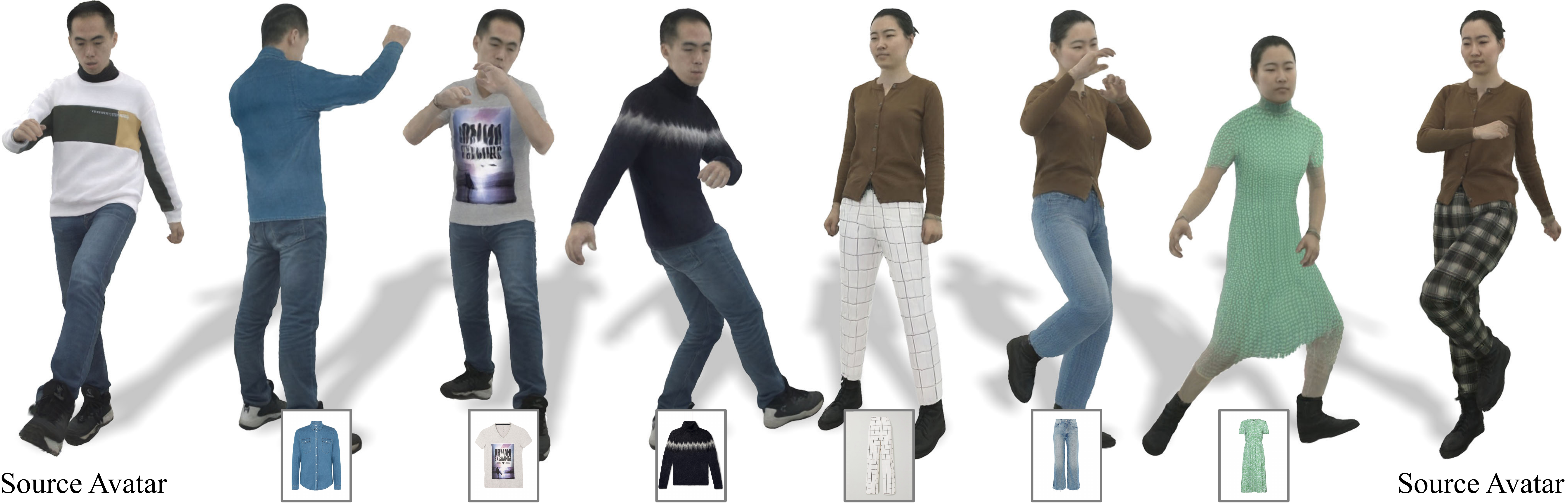}
  \captionof{figure}{We propose AvatarVTON, the first solution that enables 4D virtual try-on with free pose control, viewpoint rendering, and diverse garment selection from a single in-shop garment image.}
  \label{fig:teaser}
\end{center}
}]

\begin{abstract}
We propose AvatarVTON, the first 4D virtual try-on framework that generates realistic try-on results from a single in-shop garment image, enabling free pose control, novel-view rendering, and diverse garment choices. Unlike existing methods, AvatarVTON supports dynamic garment interactions under single-view supervision, without relying on multi-view garment captures or physics priors.
The framework consists of two key modules: (1) a Reciprocal Flow Rectifier, a prior-free optical-flow correction strategy that stabilizes avatar fitting and ensures temporal coherence; and (2) a Non-Linear Deformer, which decomposes Gaussian maps into view-pose-invariant and view-pose-specific components, enabling adaptive, non-linear garment deformations. To establish a benchmark for 4D virtual try-on, we extend existing baselines with unified modules for fair qualitative and quantitative comparisons. Extensive experiments show that AvatarVTON achieves high fidelity, diversity, and dynamic garment realism, making it well-suited for AR/VR, gaming, and digital-human applications.
\end{abstract}

\begin{IEEEkeywords}
Virtual Try-On, Animatable Avatar, 3D Gaussian Splatting
\end{IEEEkeywords}

\section{Introduction}
\label{sec:intro}
Generating high-fidelity and diverse digital humans is crucial for applications in AR/VR, gaming, and holographic communication. While recent studies reconstruct animatable avatars from multi-view, multi-pose 4D data of a single subject \cite{bagautdinov2021driving, weng2022humannerf, feng2022scarf, zheng2023avatarrex, zielonka25d3ga, qian20243dgsavatar, li2024animatable}, their diversity is limited by the need for expensive capture setups. As clothing, with its rich textures and complex topology, plays a central role in visual diversity, we advocate leveraging virtual try-on (VTON) as a scalable alternative to costly data collection, especially as 4D VTON enables controllable garment manipulation for online shopping scenarios beyond reconstruction.

3D VTON offers a promising way to enhance try-on diversity by supporting garment editing and transfer. These methods~\cite{chen2024gaussianvton, chen2024gaussianeditor, cao2024gsvton} typically utilize diffusion priors \cite{diffusion} to modify garments on 3D human bodies. However, they do not support dynamic manipulation. In contrast, animatable avatar-based garment transfer approaches \cite{feng2022scarf,lin2024layga} allow temporal clothing exchange between digital humans by disentangling garments from body geometry and reconstructing fine-grained garment details. Nevertheless, these approaches generally depend on large-scale datasets, since generating realistic textures and deformations often requires data on par with full avatar reconstruction, which limits their scalability and practical use.

Consequently, our interest shifts to the image-based VTON paradigm~\cite{han2018viton, fele2022c, xie2023gp, morelli2023ladi, yang2024d4vton, idmvton}, which requires only a single 2D garment image, thereby enabling flexible and broad clothing selection. However, extending this paradigm to 4D try-on introduces two primary challenges:

\noindent \textbf{(i) View-Pose Coupling Inconsistency.} 
Image-based VTON models lack intrinsic 3D perceptual understanding and often produce discontinuous try-on results across changing viewpoints and poses. Some methods~\cite{liu2023one23, liu2023zero123, liu2024humangaussian} improve view consistency under fixed poses, and vice versa, but they fail to generalize across diverse view-pose combinations due to entangled variations. Moreover, the scarcity of 3D datasets limits access to reliable view-pose consistency priors, underscoring the need for a strategy that enforces continuity without external supervision.

\noindent \textbf{(ii) Insufficient Non-Linearity.}
Real-world captured 3D datasets~\cite{isik2023humanrf, zheng2023avatarrex} naturally encode garment motion and enable non-linear modeling of complex deformations. In contrast, single 2D garment images inherently lack view and pose variations, making it difficult to capture realistic garment dynamics without relying on physics-based simulations.

To address these limitations, we propose \textit{AvatarVTON}, the first 4D virtual try-on framework for animatable avatars. AvatarVTON follows a character-specific paradigm, synthesizing high-fidelity try-on results from single garment images while supporting free viewpoint and pose control with extensive garment selection.
To mitigate view-pose inconsistency, we introduce a prior-free consistency enhancement strategy based on the \textit{Reciprocal Flow Rectifier (RFR)}. By leveraging flow-based corrections, RFR reduces inconsistencies during avatar fitting, ensuring coherent training and eliminating artifacts in synthesized frames.
To enable realistic garment deformations, we propose a \textit{Non-Linear Deformer} based on shared view-pose-specific Gaussian offsets. Specifically, our framework decomposes the Gaussian representation into view-pose invariant and view-pose specific components, capturing deformation dynamics beyond traditional position, rotation, and scale parameters. We then design a trifurcated structure to share deformation offsets between the source and target avatars, enabling adaptive deformation transfer while avoiding disentanglement of complex neural rendering semantics. As the first framework dedicated to 4D virtual try-on, we establish a new benchmark by extending existing baselines with additional modules for fair qualitative and quantitative analysis. Extensive experiments demonstrate that AvatarVTON achieves superior fidelity, diversity, and dynamic garment realism while maintaining high rendering quality. 

In summary, our key contributions are:
\begin{itemize}
    \item We propose AvatarVTON, a novel framework that integrates in-shop garment references with animatable avatars, enabling high-fidelity 4D virtual try-on with free viewpoint and pose control from a single garment image.
    
    \item We introduce the Reciprocal Flow Rectifier, a prior-free consistency enhancement strategy that mitigates view-pose coupling inconsistencies, ensuring coherent avatar geometry and appearance across poses and viewpoints.
    
    \item We develop the Non-Linear Deformer, a view-pose-aware Gaussian decomposition framework that introduces shared view-pose-specific Gaussian offsets, facilitating adaptive garment deformation transfer without explicit disentanglement of complex neural rendering semantics.
\end{itemize}

\section{Related work}
\label{sec:related work}
\subsection{Animatable Human Reconstruction} 
Animatable human reconstruction aims to generate pose-controllable 3D models from visual inputs. SMPL~\cite{SMPL, pavlakos2019smplx} introduces explicit parametric models with Linear Blend Skinning (LBS) for shape control. Subsequent works~\cite{saito2019pifu, huang2020arch, xiu2022icon, xiu2023econ} improve fidelity using implicit networks to predict residual nonlinear deformations. However, their reliance on low-frequency representations (\eg, occupancy fields, SDF) inherently limits the recovery of high-frequency details.
Recent advances in neural rendering~\cite{mildenhall2021nerf, kerbl20233dgs} have led to approaches~\cite{feng2022scarf, weng2022humannerf, zielonka25d3ga, qian20243dgsavatar, li2024animatable} that achieve superior visual quality. HumanNeRF~\cite{weng2022humannerf} incorporates a pose-conditioned MLP for deformation, improving realism at the cost of increased computational complexity. Animatable Gaussian~\cite{li2024animatable} introduces parametric templates and CNN-driven Gaussian generation~\cite{wang2023styleavatar}, but its entangled representations complicate editing.
These methods tightly couple the backbone network with downstream tasks, limiting adaptability and hindering further development. In contrast, we propose a decoupled framework that disentangles input and output dependencies, facilitating a more flexible and efficient try-on system.

\subsection{Image-Based Virtual Try-On}
Image-based virtual try-on has gained popularity due to its minimal input requirements and practical versatility. Building on VITON’s~\cite{han2018viton} pioneering alignment-and-synthesis paradigm, recent works~\cite{morelli2023ladi, gou2023dcivton, kim2024stableviton, idmvton, yang2024d4vton} leverage diffusion models to enhance photorealism and texture fidelity. However, relying primarily on 2D data, these methods struggle to ensure consistent results across varying viewpoints and poses. Video-based approaches, such as ViViD~\cite{fang2024vivid}, address temporal coherence by modeling frame-to-frame consistency. Nonetheless, they require continuous video input, which increases computational and memory demands, thus limiting practicality. Moreover, most existing methods depend on 2D pose priors~\cite{guler2018densepose, openpose, li2020parse}, which can introduce ambiguities when lifting to 3D representations. In contrast, our approach adopts a fully 3D pose-driven framework, enabling authentic 3D-aware virtual try-on with enhanced consistency and realism across diverse views and poses.

\subsection{3D Virtual Try-On}
3D virtual try-on extends beyond standard 3D editing by addressing garment deformation challenges unique to human motion. While text-guided editing~\cite{wang2022clipnerf, haque2023in2n, dong2023vica, igs2gs, chen2024gaussianeditor, wu2024gaussctrl} allows garment replacement, existing methods struggle with fine-grained texture specification. GaussianEditor~\cite{chen2024gaussianeditor} supports text-driven garment editing but lacks precision in controlling detailed textures. Similarly, GaussianVTON~\cite{chen2024gaussianvton} and GS-VTON~\cite{cao2024gsvton} enable single-image garment editing but lack pose awareness, limiting their applicability in pose-driven scenarios and restricting training on multi-pose datasets.
3D garment transfer methods~\cite{feng2022scarf, lin2024layga} enable clothing exchange between pre-trained digital humans by leveraging multi-view and multi-pose 3D human datasets as an implicit multi-image garment-reference paradigm. SCARF~\cite{feng2022scarf} utilizes a NeRF-mesh hybrid representation for clothing transfer, achieving commendable fidelity yet contrained by its modeling formulation. LayGA~\cite{lin2024layga} leverages Gaussian maps and layered modeling to decouple garments from animatable human representations, enabling physically plausible try-on under multi-view–pose garment datasets. However, it relies heavily on multi-view garment data and thus degenerates to its baseline, Animatable Gaussians~\cite{li2024animatable}, when only a single garment image is available. In contrast, our framework integrates the flexibility of single-image garment references with a pose-driven architecture, overcoming dataset dependency limitations while supporting dynamic, high-fidelity try-on.

\section{Method}
\label{sec:method}
\begin{figure*}[!ht]
\centering
\includegraphics[width=1.0\textwidth]{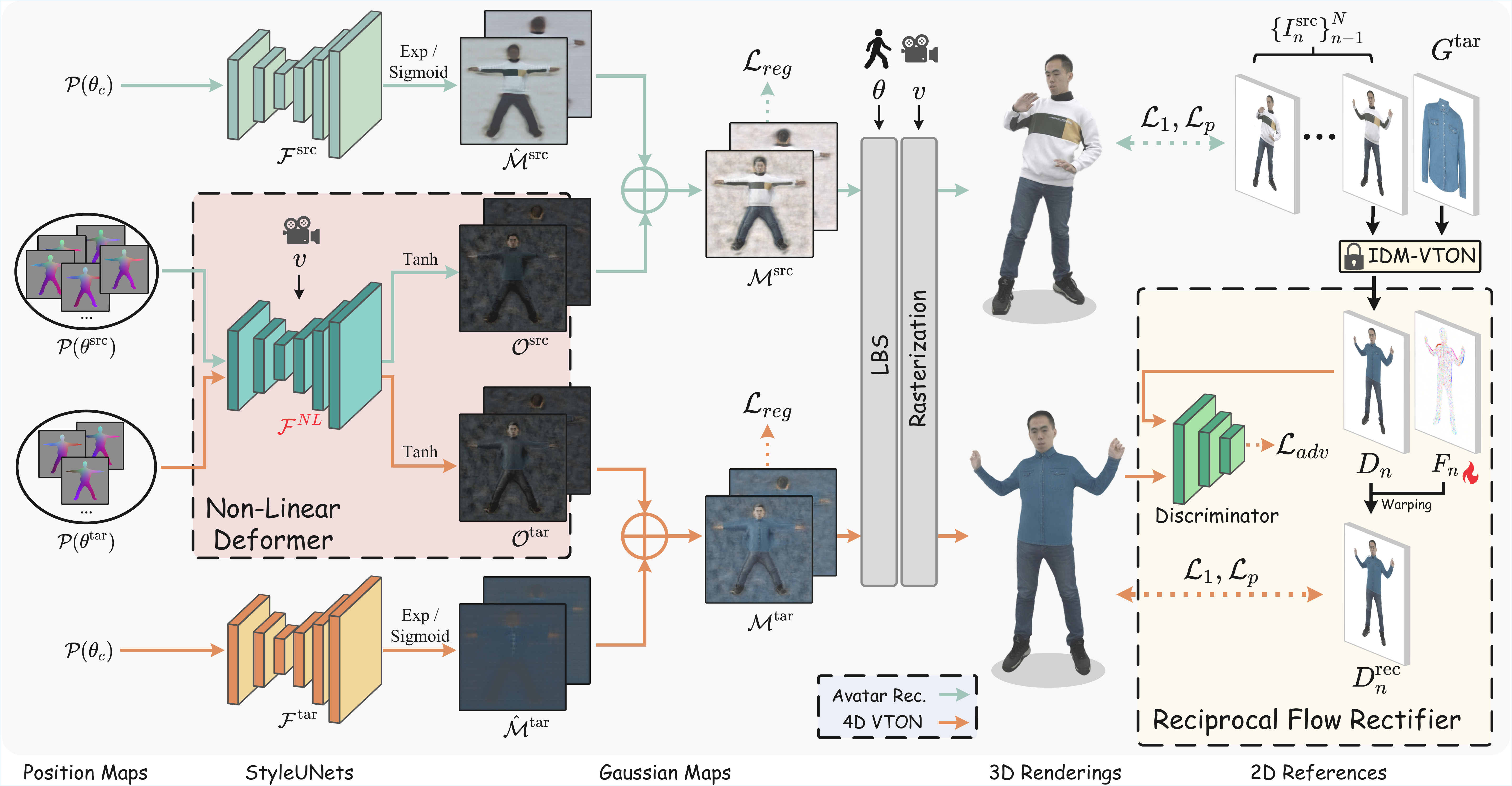}
\caption{Overview of AvatarVTON. Our framework incorporates two key techniques, namely Non-Linear Deformation Transfer, which establishes a mechanism to share nonlinear deformations across source and target tasks, and Reciprocal Flow Rectifier, which iteratively adjusts degraded frames generated by IDM-VTON~\cite{idmvton}, ensuring smooth variations during training.}
\label{fig:overview}
\end{figure*}

\subsection{Overview}
Given a set of $N$ multi-view, multi-pose images of a dressed person $\{I^{\text{src}}_n\}_{n=1}^N$ and a single in-shop garment image $G^{\text{tar}}$, our method AvatarVTON reconstructs an animatable avatar of the source subject wearing the target garment. To achieve high-quality garment transfer, we tackle the challenges of modeling complex nonlinear deformations and maintaining temporal consistency across varying poses and viewpoints.

Our framework jointly learns two tasks, animatable avatar reconstruction and 4D virtual try-on, through a parallel architecture that enables nonlinear knowledge transfer. Specifically, we design three modules based on StyleUNet~\cite{li2024animatable}: the Source StyleUNet for avatar reconstruction, the Target StyleUNet for try-on, and a Non-Linear Deformer bridging the two tasks. The first two modules transform a canonical position map into a view–pose–invariant Gaussian representation, each guided by its own objective. The Non-Linear Deformer takes a pose-driven position map and viewpoint as input, capturing view–pose–specific offsets that serve as transferable nonlinearities. The Gaussian maps and offsets are fused in canonical space, deformed via Linear Blend Skinning (LBS)~\cite{SMPL, pavlakos2019smplx}, and rendered to final results.

To enforce view–pose consistency, we generate initial 4D try-on supervision using IDM-VTON~\cite{idmvton}. However, inherent inconsistencies across views and poses in such data may lead to temporal misalignment artifacts during training. To address this, we introduce a Reciprocal Flow Rectifier that jointly optimizes optical flow and unaligned frames in a reciprocal manner, resulting in temporally coherent supervision.

These designs together allow our model to support pose-driven animation (denoted by $\theta$) and arbitrary-view rendering (denoted by $v$), producing realistic and temporally consistent results across poses and viewpoints.

\subsection{Non-Linear Deformer}
\label{3.3}
We begin by addressing the challenge of modeling nonlinear deformations of target garments in the 4D VTON task. A natural inspiration is LBS, a widely used 3D deformation method in animatable avatar modeling. However, LBS applies joint-driven linear transformations to massless points, making it insufficient for capturing complex volumetric deformations, especially with dense and expressive representations like Gaussian primitives.

Beyond algorithmic limitations, the success of learning such nonlinear deformations also depends on the richness and quality of available training data. Real human capture datasets~\cite{isik2023humanrf, zheng2023avatarrex} provide detailed observations of clothing dynamics across diverse poses and viewpoints. In avatar reconstruction, large-scale datasets (\eg, 32,000 frames) enable Gaussian maps~\cite{li2024animatable} to encode high-frequency topological variations effectively. Each pixel of a Gaussian map encodes the parameters of a Gaussian primitive, and the position map $\mathcal{P}(\theta)$ is obtained from the source subject and pose $\theta$ using Learning Parametric Templates~\cite{li2024animatable}. While effective for avatar reconstruction, this approach does not scale well to 4D VTON, as high-quality 3D supervision for garment transfer is expensive and often infeasible to acquire.

To overcome this bottleneck, we leverage 2D VTON results as alternative supervision for learning garment deformations. A naive solution would be to fine-tune the pretrained StyleUNet from avatar reconstruction using 2D VTON supervision, but this direct adaptation risks catastrophic forgetting and weakens nonlinear deformation transfer. Moreover, although 2D results are easily accessible, they lack 3D structural awareness and fail to model pose- and view-dependent dynamics. In contrast, 3D human datasets remain valuable as sources of deformation priors that can regularize 2D-guided learning and bridge the gap between 2D and 4D representations. Therefore, we decompose each Gaussian map into two components: a view–pose–invariant map synthesized by either the Source or Target StyleUNet, and a view–pose–specific offset predicted by our Non-Linear Deformer (NLD). Since pose and viewpoint changes dominate nonlinear variation, this design localizes those variations within the offset and enables deformation knowledge to be shared between avatar reconstruction and 4D VTON, while our parallel training setup prevents forgetting.

Concretely, the Source StyleUNet $\mathcal{F}^{\text{src}}(\cdot)$ and the Non-Linear Deformer $\mathcal{F}^{NL}(\cdot)$ are used for animatable avatar reconstruction, whereas the Target StyleUNet $\mathcal{F}^{\text{tar}}(\cdot)$ and the same $\mathcal{F}^{NL}(\cdot)$ are used for 4D VTON. The shared deformation mechanism is defined as
\begin{equation}
\mathcal{M}^{B} = \hat{\mathcal{M}}^{B} + \mathcal{O}^{B},
\end{equation}
\begin{equation}
\hat{\mathcal{M}}^{B} = \mathcal{F}^{B}(\mathcal{P}(\theta _{c} )), \enspace
\mathcal{O}^{B} = \mathcal{F}^{NL} (\mathcal{P}(\theta^{B} ), v),
\end{equation}
where $B\in\{\text{src}, \text{tar}\}$ indexes the branch. Here, $\hat{\mathcal{M}}^{B}$ encodes the view–pose–invariant map under the canonical pose $\theta_c$, whereas $\mathcal{O}^{B}$ models the view–pose–specific offsets of all Gaussian parameters under the driving pose $\theta^{B}$ and viewpoint $v$. While geometry-related offsets (\eg, position and scale) primarily govern spatial deformation, modeling opacity and color offsets is also crucial for preserving consistent appearance under complex illumination.

Different garments, however, exhibit distinct deformation behaviors across type and texture. This discrepancy can cause semantic inconsistencies when offsets are directly shared between branches. To mitigate this, our framework employs branch-specific activations that regulate how deformations are expressed. The reconstruction and try-on branches adopt exponential activation for scale and sigmoid activation for opacity, ensuring positive and bounded responses that preserve geometric stability and photometric consistency. In contrast, the shared branch uses a tanh activation whose symmetric and zero-centered range captures bidirectional deviations around the canonical state. This design enables the shared branch to learn generalized motion-driven deformation bases that remain independent of specific garment types, thereby enabling nonlinear knowledge from reconstruction to transfer effectively to try-on.

An additional benefit arises from how the sigmoid–tanh coupling affects opacity learning. The final opacity of each Gaussian is determined by combining the sigmoid output from task branches with the tanh modulation from the shared branch. When their combined effect approaches zero, unstable or inconsistent Gaussians are suppressed during rendering, forming a learnable soft-filtering mechanism. This adaptive behavior reduces spurious high-frequency responses along garment boundaries, mitigating the spiky artifacts commonly observed without nonlinear deformation modeling and leading to smoother geometry and improved visual coherence.

\subsection{Reciprocal Flow Rectifier}
\label{sec:rfr}
Apart from nonlinear deformation modeling, another key challenge in 2D VTON supervision lies in maintaining temporal coherence across poses and viewpoints. Directly using inconsistent 2D outputs to supervise 4D VTON can mislead the 3D model, often resulting in geometric distortions and pose-dependent texture artifacts. Although video-based VTON methods incorporate temporal cues, they still lack 3D structural awareness, and their high computational and memory costs limit practicality compared with image-based approaches.

To mitigate these inconsistencies, we adopt a two-step strategy. First, instead of using all source pose frames $\theta^{\text{src}}$ from the dataset, we uniformly sample a subset of target poses $\theta^{\text{tar}}$ (\eg, 800 frames out of approximately 32,000). This uniform sampling effectively reduces inconsistency while preserving sufficient pose and viewpoint diversity for learning. Other strategies are generally less effective: random sampling introduces instability due to the highly varied and unstructured motions in human reconstruction datasets, whereas manual curation may produce higher-quality frames but compromises robustness and generalizability.

Second, we rectify these inconsistencies using optical flow to align the 2D VTON results $\{D_n\}_{n=1}^N$. The rectified supervision $\{D^{\text{rec}}_{n}\}_{n=1}^N$ is computed as
\begin{equation}
D^{\text{rec}}_{n} = \mathrm{grid\_sample}(D_{n}, F_{n}),
D_{n} = \mathrm{VTON}(I^{\text{src}}_{n}, G^{\text{tar}}),
\end{equation}
where $\mathrm{VTON(\cdot)}$ denotes the 2D VTON model, $\{F_n\}_{n=1}^N$ represents the set of optical flows, and $\mathrm{grid\_sample}(\cdot, \cdot)$ denotes a differentiable warping operator.

The challenge lies in estimating $\{F_n\}_{n=1}^N$ without additional priors. We address this using the Target StyleUNet, which enforces pose consistency via a fixed canonical position map and fits a 3D-consistent Gaussian field. Compared to the Non-Linear Deformer whose outputs are initialized at zero, the Target StyleUNet converges faster during early training, making it the dominant avatar component. This early coherence enables it to guide the optical flow network in aligning textures from temporally degraded frames with more reliable early avatar outputs. In return, these aligned textures enhance the avatar’s texture sharpness and edge clarity as training progresses.

Geometric consistency is further maintained by employing a predefined human body model such as SMPL or a pretrained parametric template~\cite{li2024animatable}. Additionally, we use zero-order spherical harmonics to ensure view-independent Gaussian colors. Together, these constraints produce a consistent optical flow field that refines the initial 2D supervision. As a result, avatar refinement and supervision alignment form a feedback loop that reciprocally improves temporal and geometric coherence during training.

Specifically, we initialize learnable optical flows at zero and optimize them throughout training. While these flows improve view–pose consistency among training frames, they may reduce texture fidelity in the deformed regions. To counteract this, we introduce an adversarial discriminator that treats unrectified training data as real and pose-driven try-on renders as synthetic. This adversarial setup effectively restores fine texture details in the generated outputs.

\subsection{Learning Objectives}
\label{3.4}
Total objective $\mathcal{L}$ is formulated as:
\begin{equation}
\mathcal{L} = 
    \mathcal{L}_{1} + \lambda_{\text{p}} \mathcal{L}_{\text{p}} 
    + \lambda_{\text{reg}} \mathcal{L}_{\text{reg}} + \lambda_{\text{adv}} \mathcal{L}_{\text{adv}},
\label{eq:Lrender}
\end{equation}
where the $\lambda$ terms are loss weights. 

The first three losses follow standard rendering-based objectives. $\mathcal{L}_1$ is the pixel-wise L1 distance between the rendering result $\hat{x}(\theta^{\text{tar}})$ and the rectified supervision $D^{\text{rec}}_n$, $\mathcal{L}_{\text{p}}$ is the perceptual loss~\cite{zhang2018unreasonable}, and the regularization loss $\mathcal{L}_{\text{reg}} = \|\mathcal{M}^B\|_2^2$ prevents the values in Gaussian maps from growing excessively large. 

We also include an adversarial loss $\mathcal{L}_{\text{adv}}$ to facilitate adversarial training:
\begin{equation}
\begin{aligned}
    \mathcal{L}_{\text{adv}} = \mathcal{L}_{\mathrm{Dis}} + \mathcal{L}_{\mathrm{Gen}} =     
    & -\mathbb{E}_{x \sim D_{1: N}} \left[ \log \mathrm{Dis}(x) \right] \\
    & -\mathbb{E}_{\theta^{\text{tar}}} \left[ \log \left( 1 - \mathrm{Dis}(\hat{x}(\theta^{\text{tar}}) \right) \right] \\
    & -\mathbb{E}_{\theta^{\text{tar}}}[\log \mathrm{Dis}(\hat{x}(\theta^{\text{tar}}))],
\end{aligned}
\end{equation}
where $\mathrm{Dis}(\cdot)$ denotes the discriminator and $\mathrm{Gen}(\cdot)$ indicates the 4D VTON pipeline.

\section{Experiments}
\label{sec:experiment}
\subsection{Experimental Settings}

\subsubsection{Datasets}
We train our model on five subjects from two 3D human datasets: three subjects (zzr, lbn1, lbn2) from AvatarReX~\cite{zheng2023avatarrex} and two (Actor04, Actor08) from ActorsHQ~\cite{isik2023humanrf}, combined with over 50 garments from DressCode~\cite{he2024dresscode} and VITON-HD~\cite{choi2021viton}. Each subject includes 16–40 camera sequences, with around 2000 frames per sequence. We employ the image-based IDM-VTON as the 2D VTON model, and comparisons involving video-based VTON supervision are discussed in Section~\ref{sec:discussion}.

\subsubsection{Competitors}
We compare our method with state-of-the-art approaches in 2D and 3D virtual try-on, as well as extended baselines for 4D VTON. Specifically, we evaluate against 2D video-based VTON (ViViD~\cite{fang2024vivid}), 2D image-based VTON (IDM-VTON~\cite{idmvton}), and 3D editing methods (GaussianEditor~\cite{chen2024gaussianeditor}). For 4D VTON, we extend IDM-VTON with SCARF~\cite{feng2022scarf} and Animatable Gaussians (AG)~\cite{li2024animatable}, where SCARF represents a NeRF-based animatable human reconstruction baseline, and AG serves as its 3DGS-based counterpart. We further include LHM~\cite{qiu2025lhm} as a representative 4D approach capable of comparable digital human creation and animation.

\subsubsection{Metrics}  
Due to the lack of large-scale 3D human datasets with diverse garments, conventional 3D metrics such as Chamfer Distance are not applicable to quantitatively evaluate 4D VTON. As an alternative, we adopt 2D metrics computed on rendered images. Specifically, PSNR, SSIM, and LPIPS are used for reference-based evaluation by constructing paired data. For unpaired settings, we employ a CLIP-based similarity metric~\cite{radford2021clip} to measure the semantic alignment between the reference garment and the rendered result, thereby assessing texture fidelity and try-on quality. In addition, we adopt Subject Consistency (SC) and Image Quality (IQ) from VBench~\cite{huang2024vbench} to evaluate temporal coherence and per-frame image fidelity, respectively. Note that IQ is computed on individual frames and tends to favor non-4D methods that do not enforce temporal consistency. Therefore, it is excluded from the main comparison and used only in our ablation studies to analyze the contribution of different components to frame-level image quality. Finally, we also conduct a user study to assess perceptual quality, while more details on the evaluation metrics, questionnaire design, and additional analyses are provided in the supplementary materials.

\subsection{Quantitative Evaluation}
We perform paired and unpaired quantitative evaluations on three subjects from AvatarReX~\cite{zheng2023avatarrex}, each with 750–1200 frames (16–24 views $\times$ 40–50 poses), balancing preprocessing efficiency and coverage of the view–pose distribution. For paired comparisons, several fairness-promoting measures are applied, as detailed in the supplementary materials. For unpaired evaluation, competitors are tested on a dataset containing 15 garments and 10 pose sequences, totaling over 90,000 frames. Note that GaussianEditor, which requires per-frame optimization and suffers from severe temporal inconsistency, is excluded from the unpaired evaluation due to its extremely high computational cost.

\begin{table*}
\captionof{table}{Quantitative comparison with state-of-the-art approaches on AvatarReX.}
\centering

\begin{tabular}{@{}l c c c c c ccc c cc@{}}
\toprule
\multirow{2}{*}{Method} && \multirow{2}{*}{Venue} && \multirow{2}{*}{Type} && \multicolumn{3}{c}{Paired} && \multicolumn{2}{c}{Unpaired}   \\
                    &&&&&& PSNR↑ & SSIM↑ & LPIPS↓ && CLIP(\%)↑ & SC(\%)↑ \\

\cmidrule{1-1} \cmidrule{3-3} \cmidrule{5-5} \cmidrule{7-9} \cmidrule{11-12}

ViViD~\cite{fang2024vivid} 
&& ArXiv 2024
&& 2DV
&& 20.75 & 0.910 & 0.139  

&& 63.33 & 90.51
\\

IDM-VTON~\cite{idmvton} 
&& ECCV 2024
&& 2DI
&& 19.43 & 0.917 & 0.091 

&& 63.12 & 91.27
\\

GaussianEditor~\cite{chen2024gaussianeditor} 
&& CVPR 2024
&& 3D
&& 18.26 & 0.836 & 0.146
&& / & /
\\

IDM-VTON + SCARF~\cite{feng2022scarf} 
&& SIGGRAPH 2022
&& 4D
&& 13.62 & 0.837 & 0.201
&& 55.59 & 91.39
\\

IDM-VTON + AG~\cite{li2024animatable} 
&& CVPR 2024
&& 4D
&& 21.10 & 0.911 & 0.090
&& 63.22 & 91.45
\\

IDM-VTON + LHM~\cite{qiu2025lhm} 
&& ICCV 2025
&& 4D
&& 16.48 & 0.860 & 0.147
&& 63.50 & 90.86
\\

\cmidrule{1-1} \cmidrule{3-3} \cmidrule{5-5} \cmidrule{7-9} \cmidrule{11-12}
Ours 
&& /
&& 4D
&& \textbf{21.99} & \textbf{0.924} & \textbf{0.081}
&& \textbf{63.74} & \textbf{91.68}
\\
\bottomrule
\end{tabular}
\label{tab:comparison}
\end{table*}

The results in Tab.~\ref{tab:comparison} demonstrate that our method consistently outperforms all competitors in both paired and unpaired settings. Compared with ViViD, our 4D VTON framework achieves markedly better visual quality and temporal consistency, exposing the inherent limitations of 2D video-based approaches when dealing with complex human poses. Its inferior SC score primarily arises from limited input sequence length, where segmented processing degrades temporal continuity in long sequences. The paired comparison with IDM-VTON further verifies our method’s ability to mitigate inconsistency degradation commonly observed in prior models. The relatively weaker CLIP similarity of IDM-VTON is largely attributed to self-occluded poses and extreme viewpoints that fall outside its training distribution. Compared with other 4D VTON pipelines, our method yields clear improvements in rendering quality, texture fidelity, and temporal coherence.

\subsection{Qualitative Evaluation}  

\begin{figure*}
\centering
\includegraphics[width=1.0\linewidth]{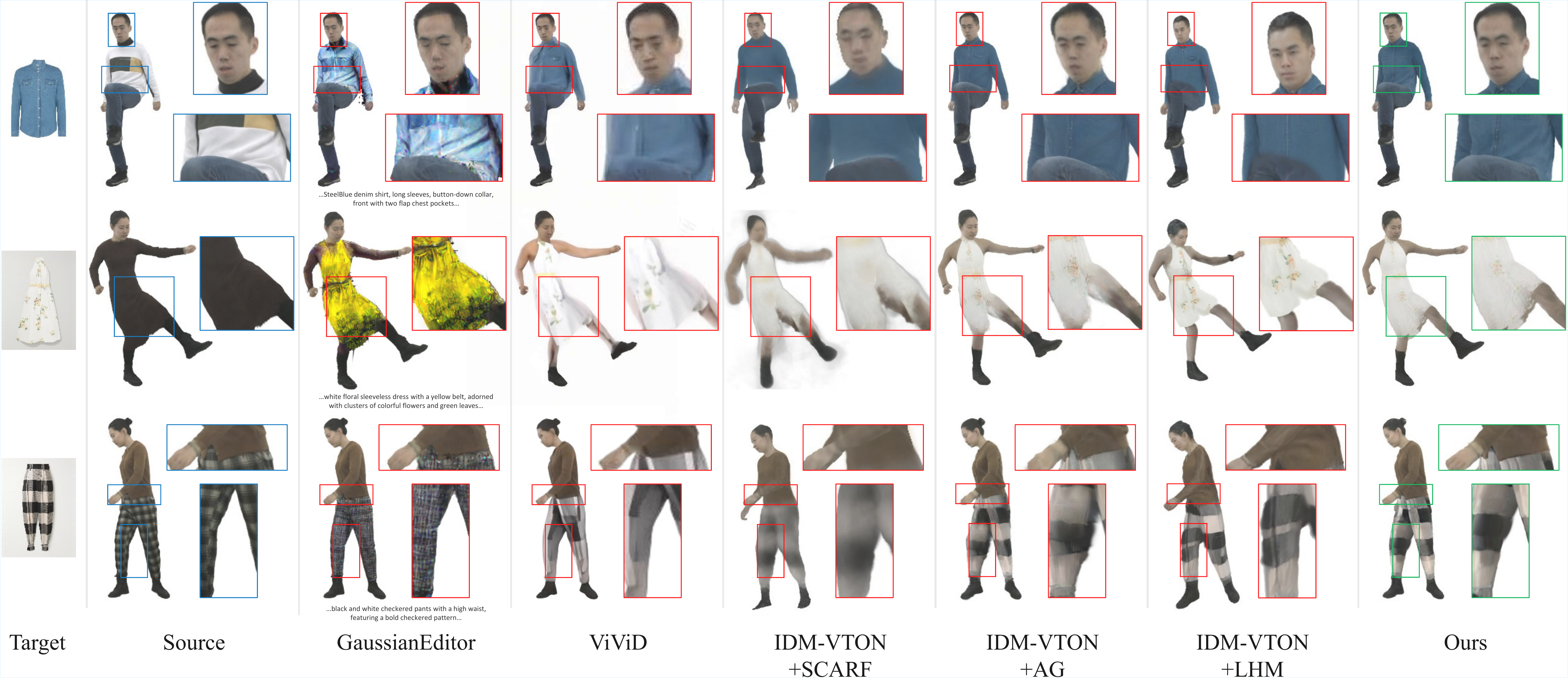}
\caption{Qualitative comparison with state-of-the-art approaches on AvatarReX~\cite{zheng2023avatarrex}.}
\label{fig:comparison}
\end{figure*}

Fig.~\ref{fig:comparison} presents qualitative comparisons under the same configuration as the quantitative evaluation. Compared with the 2D video-based baseline ViViD, our 4D VTON framework achieves substantially better temporal consistency and texture fidelity. Although ViViD provides short-term continuity through frame-level temporal modeling, it lacks explicit 3D structural reasoning, resulting in texture flickering and deformation errors under complex poses.

Among 4D baselines, IDM-VTON combined with SCARF and AG improves spatial consistency to some extent but still fails to resolve view–pose inconsistencies caused by degraded or incomplete supervision. SCARF, in particular, suffers from inaccurate pose estimation, leading to noticeable misalignment artifacts. LHM further exhibits degraded texture quality and geometric fidelity, as its one-shot formulation prevents effective utilization of multi-view supervision signals.

In contrast, our method produces the most visually consistent and realistic results, maintaining accurate geometry, plausible physical interactions, and high-fidelity textures. Note that we further incorporate AG’s generalization strategy~\cite{li2024animatable} to handle unseen view–pose combinations. Together with the demo provided in the supplementary materials, these results demonstrate the strong generalization capability of our framework.

\subsection{Ablation Study}
We conduct ablation experiments on unpaired garments to evaluate the contribution of each component. To assess the role of the Non-Linear Deformer (NLD), we remove it along with the dual-branch architecture and replace them with a single StyleUNet. For the Reciprocal Flow Rectification (RFR), we disable the reciprocal flow warping and train directly on the degraded dataset. We also compare against the baseline (IDM-VTON + AG) to examine potential regressions.
The photometric losses $\mathcal{L}_{1}$, $\mathcal{L}_{\text{p}}$, and the regularization loss $\mathcal{L}_{\text{reg}}$ are standard, so we focus validation on the adversarial loss $\mathcal{L}_{\text{adv}}$.

\begin{figure}
\centering
\includegraphics[width=0.95\linewidth]{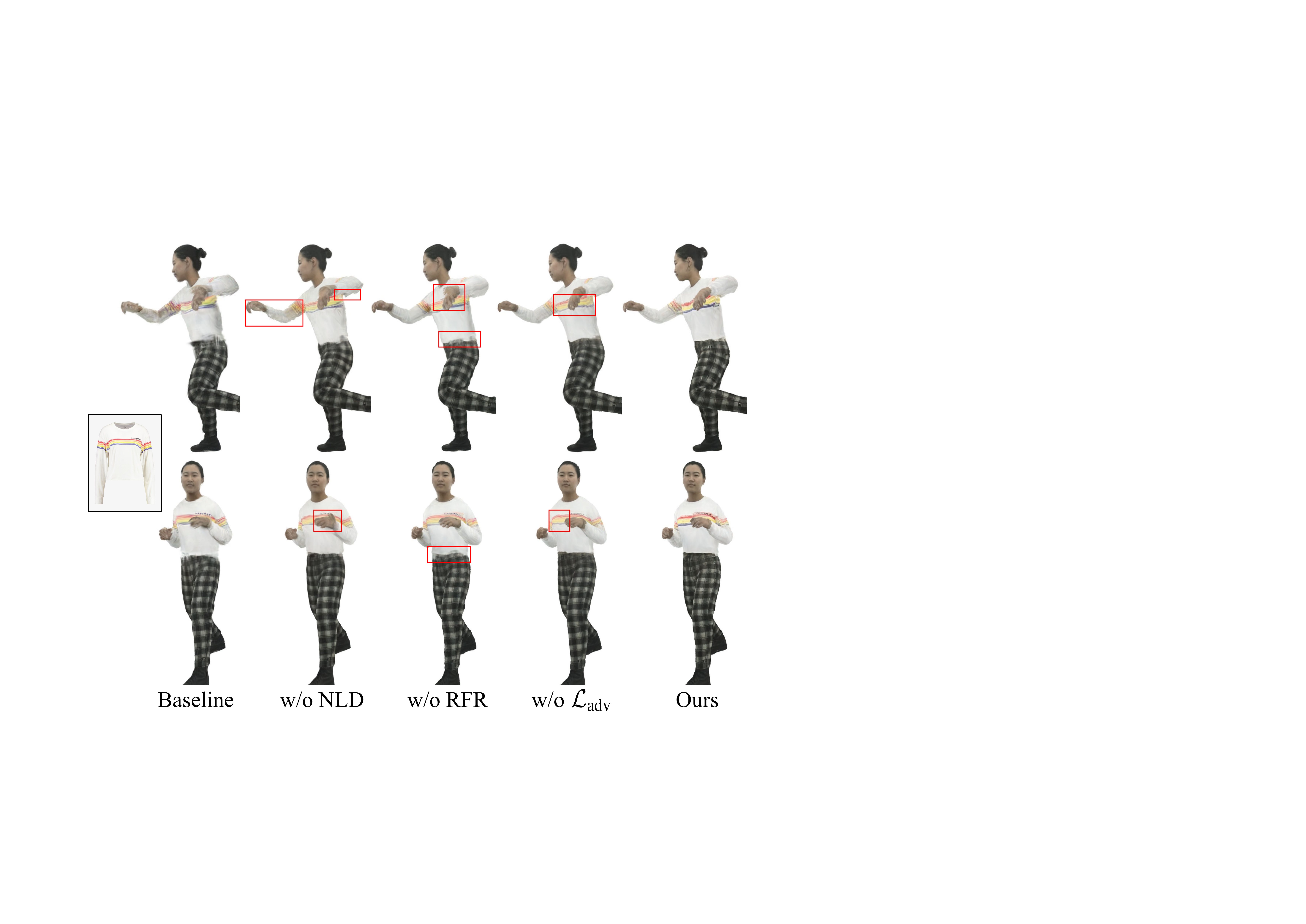}
\caption{Qualitative comparison in the ablation study.}
\label{fig:ablation}
\end{figure}

As shown in Tab.~\ref{tab:ablation} and Fig.~\ref{fig:ablation}, our complete pipeline achieves optimal or near-optimal performance on both CLIP-based and video-based metrics. The higher SC observed when excluding $\mathcal{L}_{\text{adv}}$ occurs because this metric evaluates consistency independently of garment alignment, making it insensitive to the effects of the adversarial loss.

\begin{table}
\captionof{table}{Quantitative ablation results on AvatarReX.}
\centering
\begin{tabular}{@{}l c ccc@{}}
\toprule
Variant && CLIP(\%)↑ & SC(\%)↑ & IQ(\%)↑   \\
\cmidrule{1-1} \cmidrule{3-5} 
Baseline
&& 63.22 & 91.45 & 50.07 \\

w/o NLD
&& 63.36 & 91.65 & 50.98 \\

w/o RFR
&& 63.72 & 91.51 & 53.29 \\

w/o $\mathcal L_{adv}$
&& 63.03 & 91.67 & 51.18 \\
\cmidrule{1-1} \cmidrule{3-5}
Ours
&& \textbf{63.74} & \textbf{91.68} & \textbf{53.83} \\
\bottomrule
\end{tabular}
\label{tab:ablation}
\end{table}

\subsubsection{Effectiveness of NLD}  
The decrease in IQ observed when removing NLD in Tab.~\ref{tab:ablation} indicates notable degradation in overall image quality. This deterioration primarily stems from severe geometric distortions, particularly spike-shaped artifacts on model surfaces (see Fig.~\ref{fig:ablation}), which occur when the Gaussians fail to deform coherently with the underlying geometry. In contrast, our full pipeline effectively mitigates these distortions. Moreover, NLD enhances reconstruction quality in unedited regions by producing sharper textures and clearer edges.

\begin{figure}
\centering
\includegraphics[width=\linewidth]{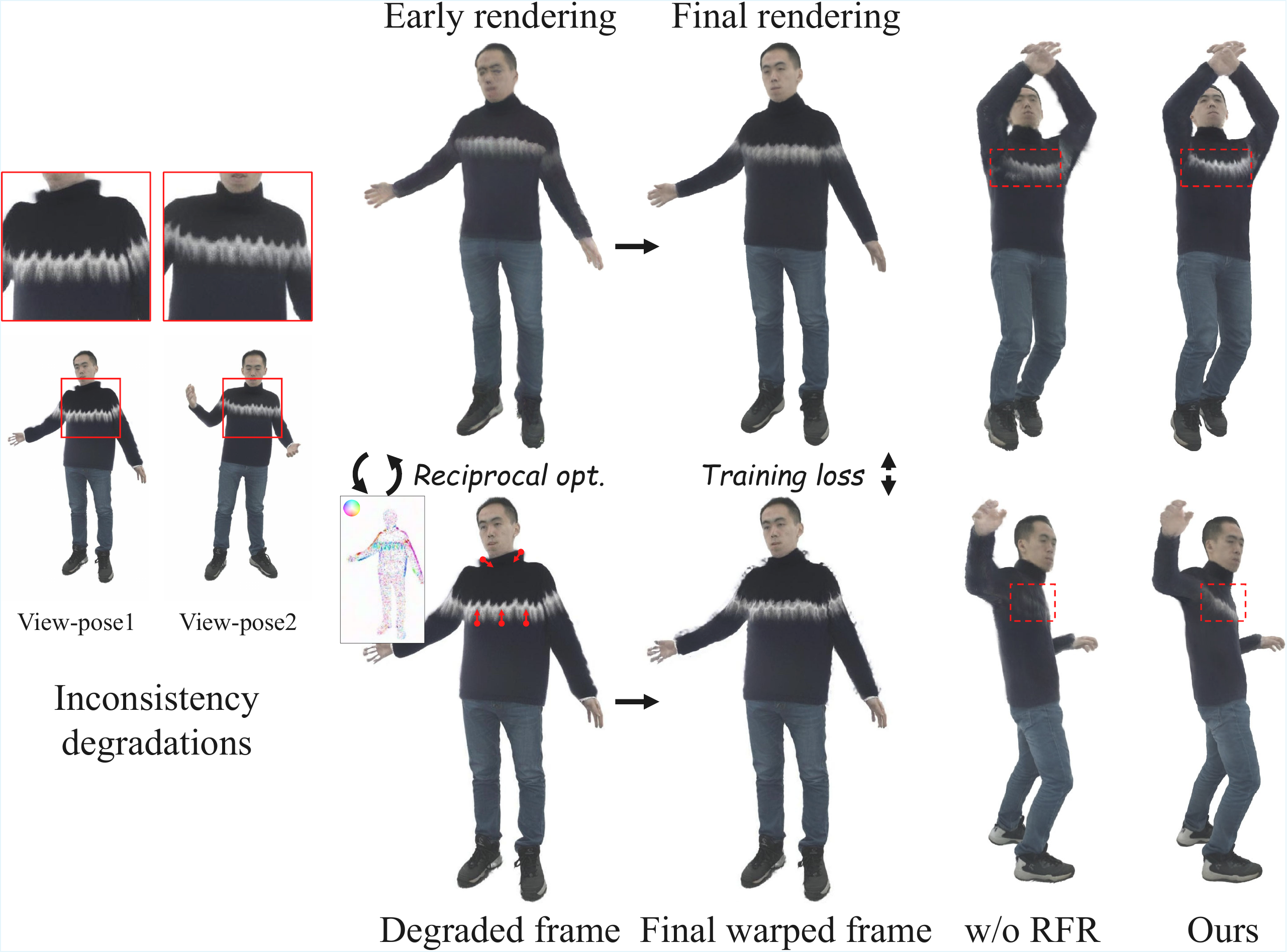}
\caption{Illustration and results of reciprocal optimization in RFR between the avatar and supervision frames, improving texture stability and cross-frame consistency.}
\label{fig:RFRvis}
\end{figure}

\subsubsection{Effectiveness of RFR}
The reduction in SC in the setting without RFR highlights its critical role in preserving temporal coherence. As shown in Fig.~\ref{fig:ablation}, removing RFR leads to blurry boundaries and distorted geometry under novel poses, while our demo further reveals flickering across view–pose transitions. In contrast, our method produces sharper textures and more stable colors. Although RFR introduces a certain trade-off in fidelity, its stability benefits generalize well to unseen view–pose scenarios, ultimately enhancing overall rendering quality and improving both CLIP similarity and visual perception metrics.

We visualize the reciprocal flow process in Fig.~\ref{fig:RFRvis}, where the flow field rectifies inconsistencies by aligning degraded supervision frames to a coherent avatar representation. This alignment resolves conflicts around boundaries and fine texture details, promoting structural continuity and stability.

\subsubsection{Effectiveness of Adversarial Loss}
As shown in Fig.~\ref{fig:ablation}, removing $\mathcal{L}_{adv}$ leads to reduced texture clarity and color accuracy due to erroneous color blending and texture jittering. The improvement in CLIP similarity reported in Tab.~\ref{tab:ablation} indicates that $\mathcal{L}_{adv}$ compensates for the fidelity degradation introduced by RFR’s trade-off between consistency and appearance fidelity, recovering fine visual details lost during consistency enforcement.

\subsection{View-Pose Coupling Consistency}
\begin{figure}
\centering
\includegraphics[width=1.0\linewidth]{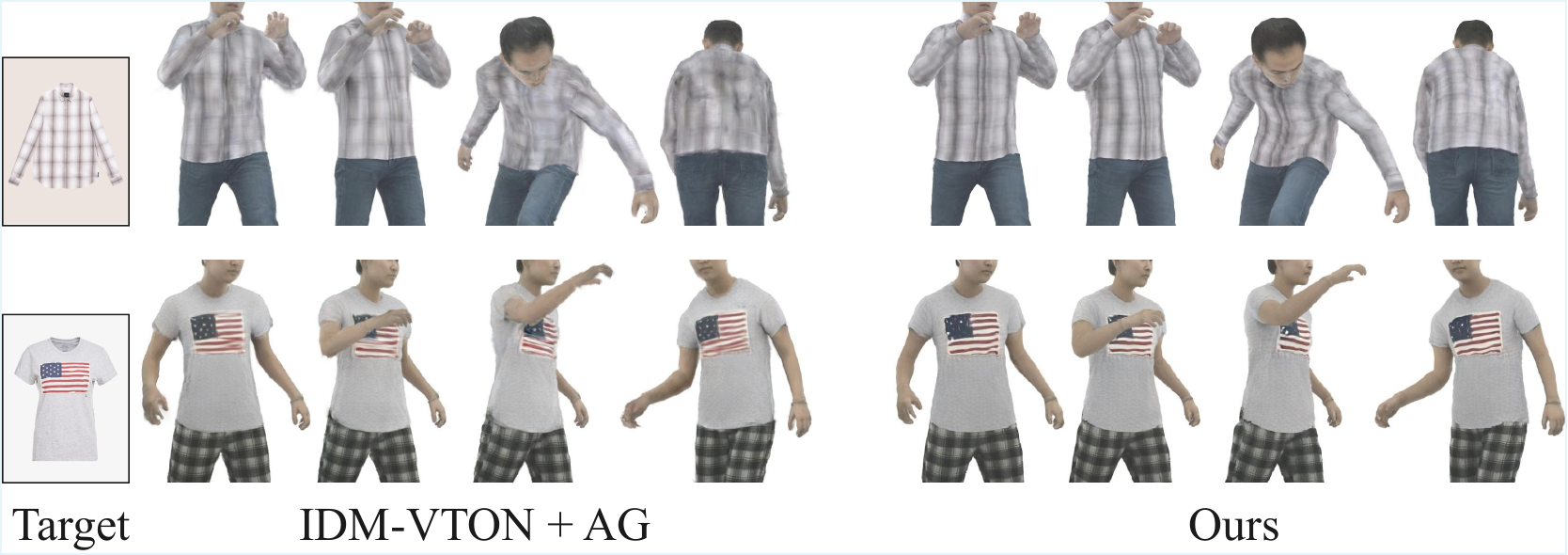}
\caption{Qualitative comparison of view–pose coupling consistency with IDM-VTON + AG.}
\label{fig:consistency}
\end{figure}

Consistency is crucial for achieving realistic 4D VTON. As illustrated in Fig.~\ref{fig:consistency}, our method maintains coherent garment deformation and stable textures across diverse viewpoints and novel poses. In contrast, the second-best 4D VTON method, IDM-VTON combined with AG, exhibits noticeable temporal flickering manifested as irregular texture patterns and inconsistent stripe counts across frames, which is also evident in the supplementary demo.

\subsection{User Study} 

\begin{figure}
\centering
\includegraphics[width=1.0\linewidth]{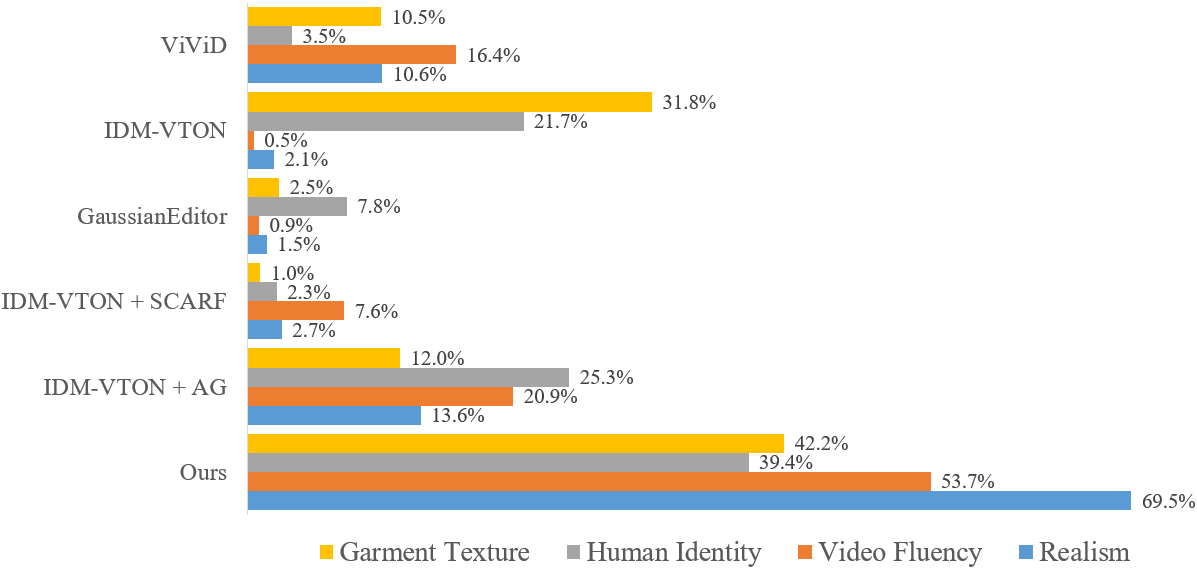}
\caption{User study results over four evaluation dimensions.}
\label{fig:userstudy_result}
\end{figure}

We conduct a user study to assess the perceptual quality of unseen garment try-on results. A total of 50 volunteers, including 25 VTON experts and 25 non-experts, complete a questionnaire consisting of 20 sets of four single-choice questions. Each question presents one human image and one garment image, along with six randomized try-on videos (our method and five competitors) rendered under identical view–pose dynamics. Participants independently rate four aspects: garment texture fidelity, human identity preservation, video temporal coherence, and overall realism.

The questionnaire covers five human subjects evenly sampled from AvatarReX~\cite{zheng2023avatarrex} and ActorsHQ~\cite{isik2023humanrf}, and 20 garments from DressCode~\cite{he2024dresscode} and VITON-HD~\cite{choi2021viton}. Garment categories are distributed in a 2:1:1 ratio among upper, lower, and dress types. The questionnaire design and representative examples are included in the supplementary materials. The results, summarized in Fig.~\ref{fig:userstudy_result}, show that our method consistently achieves higher scores than all competitors across all four evaluation dimensions.

\subsection{Cross-Type Virtual Try-On}
\begin{figure}
\centering
\includegraphics[width=0.95\linewidth]{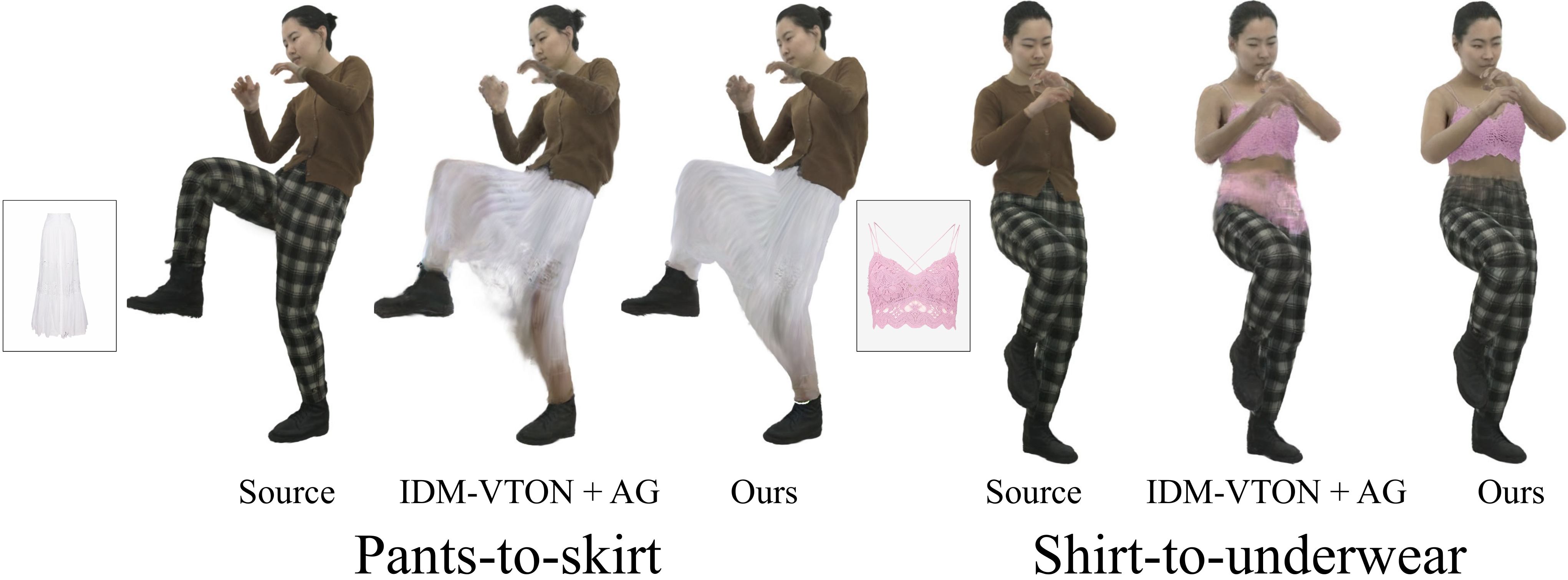}
\caption{Cross-garment-type try-on comparisons.}
\label{fig:crosstype}
\end{figure}

We showcase two representative cross-type try-on scenarios in Fig.~\ref{fig:crosstype}, including pants-to-skirt and shirt-to-underwear. These examples illustrate the challenges of under-inclusion and over-inclusion between the deformation prior of the source branch and the ideal deformation prior of the target branch. To address the geometric gap between pants and skirts, we generate a new position map for the target subject solely on degraded frames. Experimental results show that IDM-VTON combined with AG produces severe geometric and texture distortions in both scenarios, whereas our method preserves accurate geometry and high-fidelity textures. These findings demonstrate that implicit deformation transfer can effectively propagate non-trivial deformation knowledge across garment categories, validating the adaptive capability of NLD.

\subsection{Discussion, Limitations, and Future Work}\label{sec:discussion}
Among 2D supervision strategies, one might intuitively consider video-based virtual try-on models to be more suitable for 4D VTON training, as they provide explicit temporal cues that could potentially alleviate view–pose inconsistencies. However, despite these advantages, video-based methods face intrinsic limitations in 3D structural awareness, computational efficiency, and data scalability.

\begin{figure}
\centering
\includegraphics[width=0.95\linewidth]{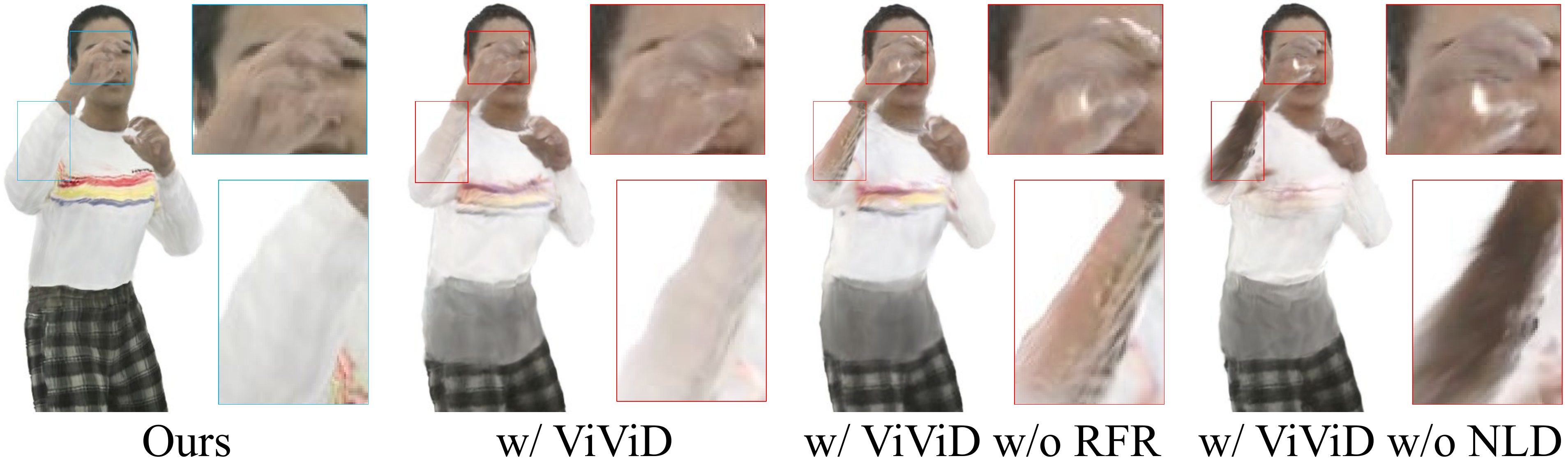}
\caption{Evaluation and ablation with ViViD supervision.}
\label{fig:vivid}
\end{figure}

Specifically, models such as ViViD incorporate temporal information across frames but remain purely 2D in formulation, lacking genuine 3D spatial reasoning. This limitation prevents them from maintaining geometric consistency across diverse viewpoints and poses. Moreover, video-based supervision requires continuous input sequences, leading to substantial computational and memory overhead. In our experiments, ViViD requires over three days of training per garment on an RTX 4090 GPU, compared to approximately three hours in our pipeline. The limited resolution imposed by GPU memory further weakens supervision quality. As shown in Fig.~\ref{fig:vivid}, ViViD produces suboptimal 4D VTON results, whereas our framework, built upon IDM-VTON with RFR, achieves a better balance among temporal coherence, detail fidelity, and computational efficiency.

\begin{figure}
\centering
\includegraphics[width=0.95\linewidth]{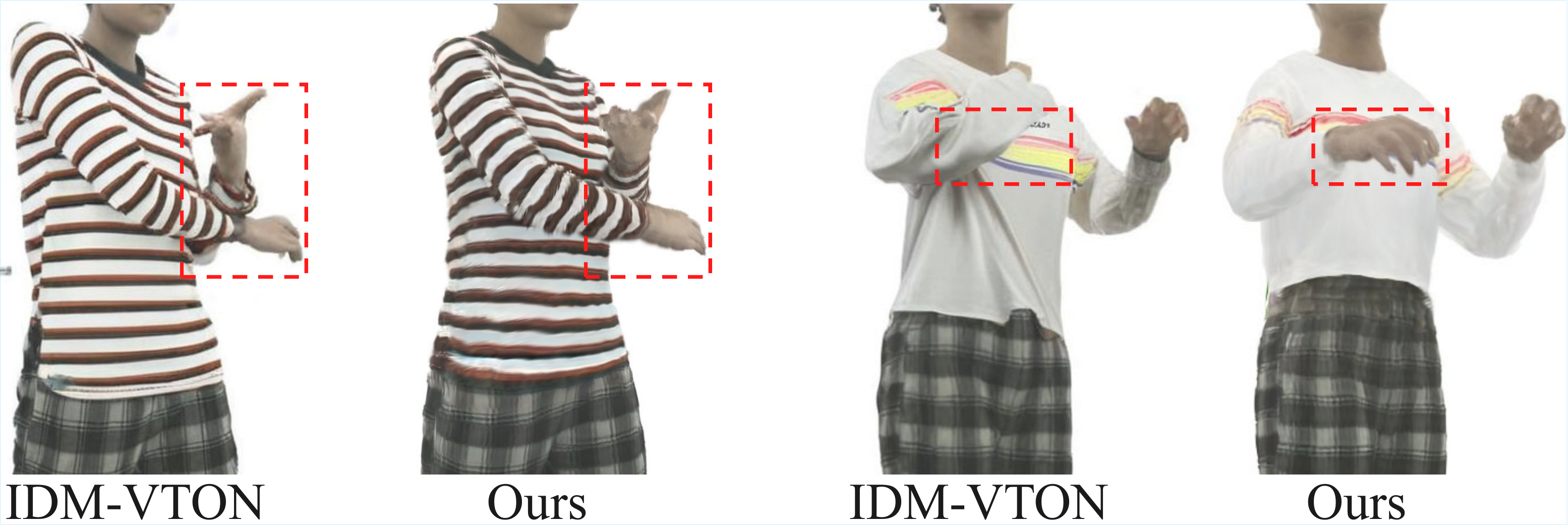}
\caption{AvatarVTON inherits the OOD limitations of existing try-on priors.}
\label{fig:limitation}
\end{figure}

While the above discussion explains the rationale behind adopting image-based supervision and the improvements it brings, certain limitations remain at the level of foundational try-on priors. Despite the advancements introduced by AvatarVTON, our framework still inherits the out-of-distribution (OOD) constraints of existing try-on priors. Because the training data of prior models seldom cover challenging scenarios such as self-occluded or extreme poses and viewpoints, artifacts may occur when rendering under unseen view–pose combinations. Although our consistency-driven design can plausibly synthesize missing regions, image quality in such cases may still be suboptimal (Fig.~\ref{fig:limitation}). Future work will focus on enhancing the 3D perception capability of the underlying prior model to better accommodate OOD view–pose variations and provide more comprehensive supervisory signals.

\section{Conclusion}
AvatarVTON introduces the first 4D virtual try-on framework utilizing single in-shop garment references, enabling dynamic pose control, multi-view rendering, and diverse garment selection without requiring multi-view datasets or physics simulations. Our solution addresses temporal instability through the Reciprocal Flow Rectifier, a prior-free optical flow correction strategy, and facilitates adaptive deformations via the Non-Linear Deformer and a pose-aware Gaussian decomposition framework. Extensive evaluations demonstrate that AvatarVTON surpasses state-of-the-art approaches in fidelity, dynamic realism, and rendering quality. We believe this will be a critical step toward high-fidelity and diverse animatable digital humans.

\bibliography{main}

\begin{thebibliography}{10}
\providecommand{\url}[1]{#1}
\csname url@samestyle\endcsname
\providecommand{\newblock}{\relax}
\providecommand{\bibinfo}[2]{#2}
\providecommand{\BIBentrySTDinterwordspacing}{\spaceskip=0pt\relax}
\providecommand{\BIBentryALTinterwordstretchfactor}{4}
\providecommand{\BIBentryALTinterwordspacing}{\spaceskip=\fontdimen2\font plus
\BIBentryALTinterwordstretchfactor\fontdimen3\font minus \fontdimen4\font\relax}
\providecommand{\BIBforeignlanguage}[2]{{%
\expandafter\ifx\csname l@#1\endcsname\relax
\typeout{** WARNING: IEEEtran.bst: No hyphenation pattern has been}%
\typeout{** loaded for the language `#1'. Using the pattern for}%
\typeout{** the default language instead.}%
\else
\language=\csname l@#1\endcsname
\fi
#2}}
\providecommand{\BIBdecl}{\relax}
\BIBdecl

\bibitem{bagautdinov2021driving}
T.~Bagautdinov, C.~Wu, T.~Simon, F.~Prada, T.~Shiratori, S.-E. Wei, W.~Xu, Y.~Sheikh, and J.~Saragih, ``Driving-signal aware full-body avatars,'' \emph{ACM TOG}, vol.~40, no.~4, pp. 1--17, 2021.

\bibitem{weng2022humannerf}
C.-Y. Weng, B.~Curless, P.~P. Srinivasan, J.~T. Barron, and I.~Kemelmacher-Shlizerman, ``Humannerf: Free-viewpoint rendering of moving people from monocular video,'' in \emph{CVPR}, 2022, pp. 16\,210--16\,220.

\bibitem{feng2022scarf}
Y.~Feng, J.~Yang, M.~Pollefeys, M.~J. Black, and T.~Bolkart, ``Capturing and animation of body and clothing from monocular video,'' in \emph{ACM SIGGRAPH}, 2022, pp. 1--9.

\bibitem{zheng2023avatarrex}
Z.~Zheng, X.~Zhao, H.~Zhang, B.~Liu, and Y.~Liu, ``Avatarrex: Real-time expressive full-body avatars,'' \emph{ACM TOG}, vol.~42, no.~4, pp. 1--19, 2023.

\bibitem{zielonka25d3ga}
W.~Zielonka, T.~Bagautdinov, S.~Saito, M.~Zollh{\"o}fer, J.~Thies, and J.~Romero, ``Drivable 3d gaussian avatars,'' in \emph{3DV}, 2025, pp. 979--990.

\bibitem{qian20243dgsavatar}
Z.~Qian, S.~Wang, M.~Mihajlovic, A.~Geiger, and S.~Tang, ``3dgs-avatar: Animatable avatars via deformable 3d gaussian splatting,'' in \emph{CVPR}, 2024, pp. 5020--5030.

\bibitem{li2024animatable}
Z.~Li, Z.~Zheng, L.~Wang, and Y.~Liu, ``Animatable gaussians: Learning pose-dependent gaussian maps for high-fidelity human avatar modeling,'' in \emph{CVPR}, 2024, pp. 19\,711--19\,722.

\bibitem{chen2024gaussianvton}
H.~Chen, Y.~Huang, H.~Huang, X.~Ge, and D.~Shao, ``Gaussianvton: 3d human virtual try-on via multi-stage gaussian splatting editing with image prompting,'' \emph{arXiv preprint arXiv:2405.07472}, 2024.

\bibitem{chen2024gaussianeditor}
Y.~Chen, Z.~Chen, C.~Zhang, F.~Wang, X.~Yang, Y.~Wang, Z.~Cai, L.~Yang, H.~Liu, and G.~Lin, ``Gaussianeditor: Swift and controllable 3d editing with gaussian splatting,'' in \emph{CVPR}, 2024, pp. 21\,476--21\,485.

\bibitem{cao2024gsvton}
Y.~Cao, M.~Hadi, L.~Pan, and Z.~Liu, ``Gs-vton: Controllable 3d virtual try-on with gaussian splatting,'' \emph{arXiv preprint arXiv:2410.05259}, 2024.

\bibitem{diffusion}
R.~Rombach, A.~Blattmann, D.~Lorenz, P.~Esser, and B.~Ommer, ``High-resolution image synthesis with latent diffusion models,'' in \emph{CVPR}, 2022, pp. 10\,684--10\,695.

\bibitem{lin2024layga}
S.~Lin, Z.~Li, Z.~Su, Z.~Zheng, H.~Zhang, and Y.~Liu, ``Layga: Layered gaussian avatars for animatable clothing transfer,'' in \emph{ACM SIGGRAPH}, 2024, pp. 1--11.

\bibitem{han2018viton}
X.~Han, Z.~Wu, Z.~Wu, R.~Yu, and L.~S. Davis, ``Viton: An image-based virtual try-on network,'' in \emph{CVPR}, 2018, pp. 7543--7552.

\bibitem{fele2022c}
B.~Fele, A.~Lampe, P.~Peer, and V.~Struc, ``C-vton: Context-driven image-based virtual try-on network,'' in \emph{WACV}, 2022, pp. 3144--3153.

\bibitem{xie2023gp}
Z.~Xie, Z.~Huang, X.~Dong, F.~Zhao, H.~Dong, X.~Zhang, F.~Zhu, and X.~Liang, ``Gp-vton: Towards general purpose virtual try-on via collaborative local-flow global-parsing learning,'' in \emph{CVPR}, 2023, pp. 23\,550--23\,559.

\bibitem{morelli2023ladi}
D.~Morelli, A.~Baldrati, G.~Cartella, M.~Cornia, M.~Bertini, and R.~Cucchiara, ``Ladi-vton: Latent diffusion textual-inversion enhanced virtual try-on,'' in \emph{ACM MM}, 2023, pp. 8580--8589.

\bibitem{yang2024d4vton}
Z.~Yang, Z.~Jiang, X.~Li, H.~Zhou, J.~Dong, H.~Zhang, and Y.~Du, ``{D}$^{4}$-vton: Dynamic semantics disentangling for differential diffusion based virtual try-on,'' in \emph{ECCV}, 2024, pp. 36--52.

\bibitem{idmvton}
Y.~Choi, S.~Kwak, K.~Lee, H.~Choi, and J.~Shin, ``Improving diffusion models for authentic virtual try-on in the wild,'' in \emph{ECCV}, 2024, pp. 206--235.

\bibitem{liu2023one23}
M.~Liu, C.~Xu, H.~Jin, L.~Chen, M.~Varma~T, Z.~Xu, and H.~Su, ``One-2-3-45: Any single image to 3d mesh in 45 seconds without per-shape optimization,'' \emph{NeurIPS}, vol.~36, pp. 22\,226--22\,246, 2023.

\bibitem{liu2023zero123}
R.~Liu, R.~Wu, B.~Van~Hoorick, P.~Tokmakov, S.~Zakharov, and C.~Vondrick, ``Zero-1-to-3: Zero-shot one image to 3d object,'' in \emph{ICCV}, 2023, pp. 9298--9309.

\bibitem{liu2024humangaussian}
X.~Liu, X.~Zhan, J.~Tang, Y.~Shan, G.~Zeng, D.~Lin, X.~Liu, and Z.~Liu, ``Humangaussian: Text-driven 3d human generation with gaussian splatting,'' in \emph{CVPR}, 2024, pp. 6646--6657.

\bibitem{isik2023humanrf}
M.~I\c{s}{\i}k, M.~Rünz, M.~Georgopoulos, T.~Khakhulin, J.~Starck, L.~Agapito, and M.~Nießner, ``Humanrf: High-fidelity neural radiance fields for humans in motion,'' \emph{ACM TOG}, vol.~42, no.~4, pp. 1--12, 2023.

\bibitem{SMPL}
M.~Loper, N.~Mahmood, J.~Romero, G.~Pons-Moll, and M.~J. Black, ``{SMPL}: A skinned multi-person linear model,'' \emph{ACM TOG}, vol.~34, no.~6, pp. 248:1--248:16, 2015.

\bibitem{pavlakos2019smplx}
G.~Pavlakos, V.~Choutas, N.~Ghorbani, T.~Bolkart, A.~A. Osman, D.~Tzionas, and M.~J. Black, ``Expressive body capture: 3d hands, face, and body from a single image,'' in \emph{CVPR}, 2019, pp. 10\,975--10\,985.

\bibitem{saito2019pifu}
S.~Saito, Z.~Huang, R.~Natsume, S.~Morishima, A.~Kanazawa, and H.~Li, ``Pifu: Pixel-aligned implicit function for high-resolution clothed human digitization,'' in \emph{ICCV}, 2019, pp. 2304--2314.

\bibitem{huang2020arch}
Z.~Huang, Y.~Xu, C.~Lassner, H.~Li, and T.~Tung, ``Arch: Animatable reconstruction of clothed humans,'' in \emph{CVPR}, 2020, pp. 3093--3102.

\bibitem{xiu2022icon}
Y.~Xiu, J.~Yang, D.~Tzionas, and M.~J. Black, ``Icon: Implicit clothed humans obtained from normals,'' in \emph{CVPR}, 2022, pp. 13\,286--13\,296.

\bibitem{xiu2023econ}
Y.~Xiu, J.~Yang, X.~Cao, D.~Tzionas, and M.~J. Black, ``Econ: Explicit clothed humans optimized via normal integration,'' in \emph{CVPR}, 2023, pp. 512--523.

\bibitem{mildenhall2021nerf}
B.~Mildenhall, P.~P. Srinivasan, M.~Tancik, J.~T. Barron, R.~Ramamoorthi, and R.~Ng, ``Nerf: Representing scenes as neural radiance fields for view synthesis,'' in \emph{ECCV}, 2020.

\bibitem{kerbl20233dgs}
B.~Kerbl, G.~Kopanas, T.~Leimk{\"u}hler, and G.~Drettakis, ``3d gaussian splatting for real-time radiance field rendering.'' \emph{ACM TOG}, vol.~42, no.~4, pp. 139--1, 2023.

\bibitem{wang2023styleavatar}
L.~Wang, X.~Zhao, J.~Sun, Y.~Zhang, H.~Zhang, T.~Yu, and Y.~Liu, ``Styleavatar: Real-time photo-realistic portrait avatar from a single video,'' in \emph{ACM SIGGRAPH}, 2023, pp. 1--10.

\bibitem{gou2023dcivton}
J.~Gou, S.~Sun, J.~Zhang, J.~Si, C.~Qian, and L.~Zhang, ``Taming the power of diffusion models for high-quality virtual try-on with appearance flow,'' in \emph{ACM MM}, 2023, pp. 7599--7607.

\bibitem{kim2024stableviton}
J.~Kim, G.~Gu, M.~Park, S.~Park, and J.~Choo, ``Stableviton: Learning semantic correspondence with latent diffusion model for virtual try-on,'' in \emph{CVPR}, 2024, pp. 8176--8185.

\bibitem{fang2024vivid}
Z.~Fang, W.~Zhai, A.~Su, H.~Song, K.~Zhu, M.~Wang, Y.~Chen, Z.~Liu, Y.~Cao, and Z.-J. Zha, ``Vivid: Video virtual try-on using diffusion models,'' \emph{arXiv preprint arXiv:2405.11794}, 2024.

\bibitem{guler2018densepose}
R.~A. G{\"u}ler, N.~Neverova, and I.~Kokkinos, ``Densepose: Dense human pose estimation in the wild,'' in \emph{CVPR}, 2018, pp. 7297--7306.

\bibitem{openpose}
Z.~Cao, G.~Hidalgo, T.~Simon, S.-E. Wei, and Y.~Sheikh, ``Openpose: Realtime multi-person 2d pose estimation using part affinity fields,'' \emph{IEEE TPAMI}, vol.~43, no.~1, pp. 172--186, 2019.

\bibitem{li2020parse}
P.~Li, Y.~Xu, Y.~Wei, and Y.~Yang, ``Self-correction for human parsing,'' \emph{IEEE TPAMI}, vol.~44, no.~6, pp. 3260--3271, 2020.

\bibitem{wang2022clipnerf}
C.~Wang, M.~Chai, M.~He, D.~Chen, and J.~Liao, ``Clip-nerf: Text-and-image driven manipulation of neural radiance fields,'' in \emph{CVPR}, 2022, pp. 3835--3844.

\bibitem{haque2023in2n}
A.~Haque, M.~Tancik, A.~A. Efros, A.~Holynski, and A.~Kanazawa, ``Instruct-nerf2nerf: Editing 3d scenes with instructions,'' in \emph{ICCV}, 2023, pp. 19\,740--19\,750.

\bibitem{dong2023vica}
J.~Dong and Y.-X. Wang, ``Vica-nerf: View-consistency-aware 3d editing of neural radiance fields,'' \emph{NeurIPS}, vol.~36, pp. 61\,466--61\,477, 2023.

\bibitem{igs2gs}
\BIBentryALTinterwordspacing
C.~Vachha and A.~Haque, ``Instruct-gs2gs: Editing 3d gaussian splats with instructions,'' 2024. [Online]. Available: \url{https://instruct-gs2gs.github.io/}
\BIBentrySTDinterwordspacing

\bibitem{wu2024gaussctrl}
J.~Wu, J.-W. Bian, X.~Li, G.~Wang, I.~Reid, P.~Torr, and V.~A. Prisacariu, ``Gaussctrl: Multi-view consistent text-driven 3d gaussian splatting editing,'' in \emph{ECCV}, 2024, pp. 55--71.

\bibitem{zhang2018unreasonable}
R.~Zhang, P.~Isola, A.~A. Efros, E.~Shechtman, and O.~Wang, ``The unreasonable effectiveness of deep features as a perceptual metric,'' in \emph{CVPR}, 2018, pp. 586--595.

\bibitem{he2024dresscode}
K.~He, K.~Yao, Q.~Zhang, J.~Yu, L.~Liu, and L.~Xu, ``Dresscode: Autoregressively sewing and generating garments from text guidance,'' \emph{ACM TOG}, vol.~43, no.~4, pp. 1--13, 2024.

\bibitem{choi2021viton}
S.~Choi, S.~Park, M.~Lee, and J.~Choo, ``Viton-hd: High-resolution virtual try-on via misalignment-aware normalization,'' in \emph{CVPR}, 2021, pp. 14\,131--14\,140.

\bibitem{qiu2025lhm}
L.~Qiu, X.~Gu, P.~Li, Q.~Zuo, W.~Shen, J.~Zhang, K.~Qiu, W.~Yuan, G.~Chen, Z.~Dong \emph{et~al.}, ``Lhm: Large animatable human reconstruction model from a single image in seconds,'' \emph{arXiv preprint arXiv:2503.10625}, 2025.

\bibitem{radford2021clip}
A.~Radford, J.~W. Kim, C.~Hallacy, A.~Ramesh, G.~Goh, S.~Agarwal, G.~Sastry, A.~Askell, P.~Mishkin, J.~Clark \emph{et~al.}, ``Learning transferable visual models from natural language supervision,'' in \emph{ICML}, 2021, pp. 8748--8763.

\bibitem{huang2024vbench}
Z.~Huang, Y.~He, J.~Yu, F.~Zhang, C.~Si, Y.~Jiang, Y.~Zhang, T.~Wu, Q.~Jin, N.~Chanpaisit \emph{et~al.}, ``Vbench: Comprehensive benchmark suite for video generative models,'' in \emph{CVPR}, 2024, pp. 21\,807--21\,818.

\end{thebibliography}
\bibliographystyle{IEEEtran}

\newpage
\vfill

\end{document}